\documentclass{article}
\usepackage{spconf,amsmath,amssymb,amsfonts,graphicx,booktabs,cite,url,microtype,balance}

\title{BEYOND RECONSTRUCTION ERROR: ANALYTICAL AND DATA-DRIVEN ACTION TOKENIZATION FOR AUTOREGRESSIVE VISION-LANGUAGE-ACTION MODELS}

\name{Yuxin Yang$^{1,\dagger}$, Gaohan He$^{2,\dagger}$, Changxue Guan$^{1}$, Hangming Liu$^{3,*}$\thanks{$^\dagger$Equal contribution. $^*$Corresponding author: Hangming Liu.}}
\address{$^{1}$College of Artificial Intelligence, Southwest University \qquad $^{2}$Ocean College, Zhejiang University\\
$^{3}$Tianfu Securities Co., Ltd.\\
\small tianyiyoung127@gmail.com \quad Hannah.he@zju.edu.cn \quad gcx\_0601@qq.com \quad liuhm@tianfuzq.cn}

\makeatletter
\def\@maketitle{\newpage
 \null
 \vskip 0.8em \begin{center}
 {\large \bf \@title \par} \vskip 0.8em {\large \lineskip .5em
\begin{tabular}[t]{c}\@name \\ \@address
 \end{tabular}\par} \end{center}
 \par
 \vskip 0.8em}
\makeatother

\begin{document}

\maketitle

\begin{abstract}
Discrete action tokenization is central to autoregressive vision-language-action (VLA) models, yet action representations are often evaluated primarily through reconstruction fidelity. We ask which representation properties actually matter for closed-loop control by comparing fixed analytical, data-driven linear, and nonlinear neural representations under a unified tokenization interface. Across rate--distortion analysis, sequence-modeling diagnostics, and 3,500 LIBERO rollouts, representation rankings change with the evaluation criterion. PCA achieves lower nominal reconstruction error than Temporal-DCT, but produces less predictable token sequences and 3.0 percentage points lower mean seen-task success across three policy-training seeds, with the policy ordering reversing in one seed. In a matched seed-42 ablation, an autoencoder further reduces reconstruction error yet does not yield the strongest policy and exhibits greater sensitivity to discrete token perturbations. These findings show that reconstruction fidelity alone cannot reliably select action representations for autoregressive control, motivating joint evaluation of geometric fidelity, sequence predictability, decoder stability, and closed-loop performance.
\end{abstract}


\enlargethispage{\baselineskip}
\section{Introduction}
\label{sec:intro}

Autoregressive vision-language-action (VLA) models formulate robot control as sequence prediction by mapping continuous actions to discrete tokens~\cite{rt2,openvla,fast,faster,moeactok}. This interface is consequential: an action tokenizer must preserve task-relevant motion information while producing sequences that remain compact and predictable for autoregressive decoding. Consequently, tokenizer quality can directly affect closed-loop execution rather than merely compression efficiency. Early VLA systems rely on per-dimension discretization~\cite{rt2,openvla}, while recent approaches increasingly tokenize multi-step action chunks to reduce sequence length and capture temporal structure.

FAST~\cite{fast}, for example, uses a fixed temporal Discrete Cosine Transform (DCT)~\cite{ahmed1974dct} to exploit trajectory smoothness, whereas subsequent approaches explore learned vector-quantized codebooks~\cite{faster} and mixture-of-experts tokenizers~\cite{moeactok}. Despite these architectural advances, a basic representation question remains unresolved. Action tokenizers are often characterized by compression or reconstruction quality, yet autoregressive control imposes additional requirements: tokens must remain predictable to the policy, and discrete prediction errors should not induce disproportionately large changes in decoded actions. Thus, lower geometric reconstruction error does not necessarily imply better closed-loop control. This raises a fundamental question: \textbf{what properties of an action representation actually matter for autoregressive robotic execution?} This suggests viewing action tokenization not merely as a compression problem, but as an interface jointly coupling geometric representation, autoregressive sequence modeling, and error propagation into physical actions.

\begin{figure*}[t]
\centering
\includegraphics[width=0.91\textwidth]{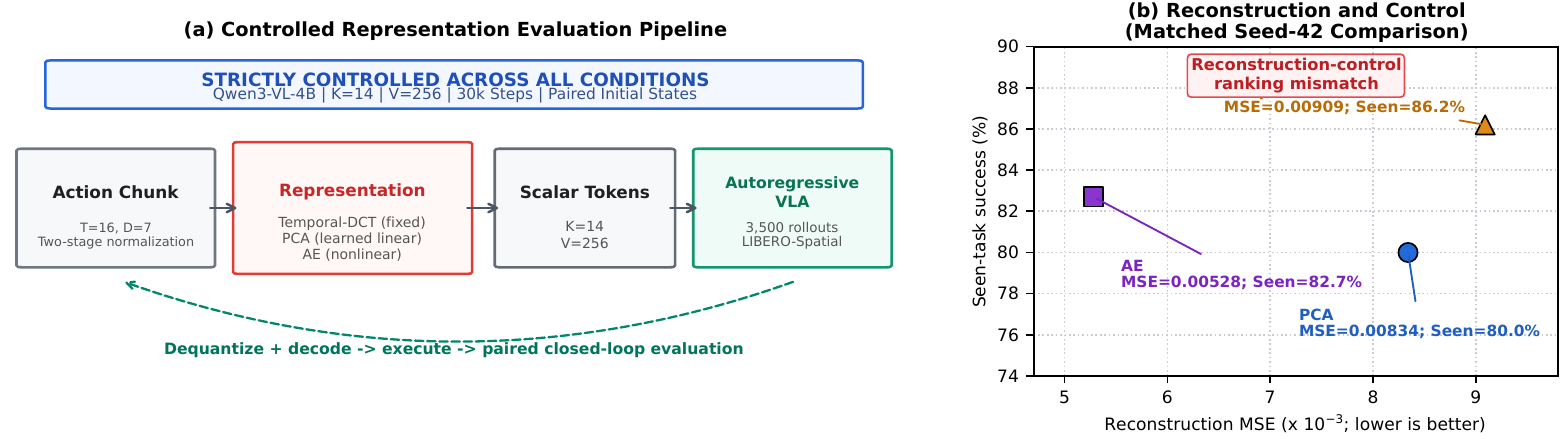}
\vspace{-3mm}
\caption{(a) Controlled evaluation pipeline: only the action representation is varied, while the VLA backbone, token budget, vocabulary, optimizer, and rollout protocol are matched. (b) Seed-42 reconstruction--control misalignment: the representation with the lowest reconstruction MSE is not the one with the highest seen-task success.}
\label{fig:pipeline_tradeoff}
\vspace{-4mm}
\end{figure*}

We investigate this question through three representative representation families under a unified tokenization interface (Fig.~\ref{fig:pipeline_tradeoff}). Temporal-DCT provides a fixed analytical harmonic prior; Principal Component Analysis (PCA) provides a data-driven orthogonal basis that captures cross-axis correlations; and an MLP autoencoder provides a flexible nonlinear latent representation. By keeping normalization, quantization, token budget, VLA backbone, optimization, and rollout protocols matched within each comparison, we isolate how representation choice affects rate-distortion behavior, sequence predictability, quantization sensitivity, and closed-loop performance.

Our main findings are threefold:
\begin{itemize}\setlength{\itemsep}{0pt}\setlength{\parsep}{0pt}\setlength{\parskip}{0pt}\setlength{\topsep}{1.5pt}
    \item \textbf{Representation Trade-offs}: Across 20 rate--distortion regimes and cross-distribution tests, PCA attains lower nominal reconstruction error than Temporal-DCT, while DCT is more robust under coarse quantization and yields more predictable sequences.
    \item \textbf{Closed-Loop Policy Performance}: In 3,500 closed-loop rollouts with Qwen3-VL-4B on LIBERO-Spatial, Temporal-DCT exceeds PCA by 3.0 pp on average across three seeds, despite its higher nominal reconstruction error; the ordering reverses in one seed.
    \item \textbf{Reconstruction--Control Misalignment}: In a matched ablation, an autoencoder attains the lowest reconstruction error but not the highest success rate. Its larger perturbation gain further identifies decoder stability as a property hidden by reconstruction MSE.
\end{itemize}

\enlargethispage{3\baselineskip}
\section{Basis Action Tokenization Pipeline}
\label{sec:formulation}
To isolate the effect of the underlying representation basis, we deliberately omit downstream BPE compression and expose all methods through an identical fixed-$K$ scalar-token interface. Thus, our Temporal-DCT condition evaluates the analytical representation stage underlying FAST~\cite{fast} rather than reproducing the complete FAST tokenizer.

\begin{table}[t]
\centering
\small
\caption{Reconstruction MSE across 20 rate-distortion regimes on LIBERO-Spatial. P/D denotes PCA/Temporal-DCT as the lower-MSE representation; values indicate relative MSE reduction $\Delta$ (\%). Bold highlights nominal setting.}
\label{tab:e11_sweep}
\vspace{1.5mm}
\begin{tabular}{cccccc}
\toprule
\textbf{Tokens $K$} & \textbf{2-Bit} & \textbf{4-Bit} & \textbf{6-Bit} & \textbf{8-Bit} & \textbf{16-Bit} \\
\midrule
$K=7$  & D 13.3 & P 20.7 & P 24.9 & P 25.2         & P 25.2 \\
$K=14$ & D 10.2 & P 3.0  & P 7.7  & \textbf{P 8.3} & P 8.3  \\
$K=21$ & D 9.4  & D 1.8  & P 1.6  & P 2.5          & P 2.5  \\
$K=28$ & D 9.1  & D 1.7  & P 2.8  & P 4.5          & P 4.4  \\
\bottomrule
\end{tabular}
\vspace{-3.5mm}
\end{table}

\begin{table*}[t]
\centering
\caption{Closed-loop success (\%) on LIBERO-Spatial over three seeds (50 paired initial states/task). Tasks 0--8 are seen; all six PCA/DCT policies score 0/50 on held-out Task~9. $\pm$ denotes standard deviation across policy-training seeds.}
\label{tab:main_rollout}
\vspace{1.5mm}
\setlength{\tabcolsep}{4.2pt}
\resizebox{0.84\textwidth}{!}{
\begin{tabular}{lcccccccccc}
\toprule
\textbf{Condition} & \textbf{Task 0} & \textbf{Task 1} & \textbf{Task 2} & \textbf{Task 3} & \textbf{Task 4} & \textbf{Task 5} & \textbf{Task 6} & \textbf{Task 7} & \textbf{Task 8} & \textbf{Seen Avg (0--8)} \\
\midrule
PCA (seed 42)  & 96\% & 90\% & 94\% & 66\% & 44\% & 88\% & 94\% & 92\% & 56\% & 80.0\% (360/450) \\
PCA (seed 43)  & 88\% & 84\% & 100\% & 70\% & 78\% & 92\% & 96\% & 90\% & 70\% & 85.3\% (384/450) \\
PCA (seed 44)  & 94\% & 86\% & 100\% & 80\% & 60\% & 84\% & 92\% & 86\% & 76\% & 84.2\% (379/450) \\
\midrule
\textbf{PCA (3-Seed Mean)} & 92.7\% & 86.7\% & \textbf{98.0\%} & 72.0\% & \textbf{60.7\%} & 88.0\% & \textbf{94.0\%} & \textbf{89.3\%} & 67.3\% & 83.2\% $\pm$ 2.8\% \\
\midrule
DCT (seed 42)  & 90\% & 88\% & 96\% & 84\% & 56\% & 90\% & 94\% & 84\% & 94\% & 86.2\% (388/450) \\
DCT (seed 43)  & 90\% & 96\% & 98\% & 92\% & 82\% & 90\% & 98\% & 88\% & 82\% & 90.7\% (408/450) \\
DCT (seed 44)  & 92\% & 96\% & 86\% & 94\% & 26\% & 86\% & 88\% & 86\% & 80\% & 81.6\% (367/450) \\
\midrule
\textbf{DCT (3-Seed Mean)} & 90.7\% & \textbf{93.3\%} & 93.3\% & \textbf{90.0\%} & 54.7\% & \textbf{88.7\%} & 93.3\% & 86.0\% & \textbf{85.3\%} & \textbf{86.2\%} $\pm$ 4.6\% \\
\midrule
\textbf{$\Delta$ (PCA $-$ DCT)} & +2.0 & $-$6.7 & +4.7 & $-$18.0 & +6.0 & $-$0.7 & +0.7 & +3.3 & $-$18.0 & \textbf{$-$3.0 pp} \\
\bottomrule
\end{tabular}
}
\vspace{-3.5mm}
\end{table*}

\subsection{Trajectory Chunking and Normalization}
Following action chunking~\cite{act,diffusion_policy,fast}, continuous control trajectories are sliced into temporal windows of horizon $T=16$ with action dimension $D=7$ (3D Cartesian translation, 3D axis-angle rotation, and 1D binary gripper state). A single chunk is flattened into $\mathbf{a} \in \mathbb{R}^{M}$, where $M = T \times D = 112$.

To guarantee strictly leakage-free conditioning, normalizers are fitted exclusively on training demonstrations (Tasks 0--8): (1)~\emph{Coordinate Scaling}: 6-DoF end-effector pose commands are scaled to $[-1, 1]$ based on training bounds, while binary gripper commands are kept unscaled in $[0, 1]$; (2)~\emph{Standardization}: the flattened $112\text{D}$ vector is $z$-score standardized as $\tilde{\mathbf{a}} = (\mathbf{a} - \boldsymbol{\mu}) \oslash \boldsymbol{\sigma}$.

\subsection{Analytical, Linear, and Neural Tokenizers}
A linear tokenizer projects standardized chunk $\tilde{\mathbf{a}} \in \mathbb{R}^{M}$ onto $K$ orthonormal basis vectors $\{\mathbf{b}_k\}_{k=1}^K$ ($\mathbf{B} \in \mathbb{R}^{K \times M}$, $\mathbf{B} \mathbf{B}^\top = \mathbf{I}_K$):
\begin{equation}
\setlength{\abovedisplayskip}{2.5pt}\setlength{\belowdisplayskip}{2.5pt}
    \mathbf{c} = \mathbf{B} \tilde{\mathbf{a}} \in \mathbb{R}^K, \quad \hat{\mathbf{a}} = \mathbf{B}^\top \mathbf{c} \in \mathbb{R}^M.
    \label{eq:projection}
\end{equation}

\noindent\textbf{Temporal-DCT Basis}: Following FAST~\cite{fast}, independent 1D DCT-II transforms are applied along the time horizon $T$ for each actuator channel. The 1D DCT matrix $\mathbf{C} \in \mathbb{R}^{T \times T}$ is defined by:
\begin{equation}
\setlength{\abovedisplayskip}{2.5pt}\setlength{\belowdisplayskip}{2.5pt}
    C_{u, t} = \sqrt{\frac{2}{T}} \alpha_u \cos\left( \frac{\pi (2t + 1) u}{2T} \right),
\end{equation}
where $\alpha_0 = 1/\sqrt{2}$ and $\alpha_u = 1$ for $u \ge 1$. To ensure all channels remain controllable, basis vectors are assigned in frequency priority: for $K=7m$, we retain the first $m$ frequency components across all $D=7$ channels ($K\in\{7,14,21,28\}$ corresponds to $m\in\{1,2,3,4\}$ coefficients/channel, with nominal $K=14$, $m=2$).

\noindent\textbf{PCA / Karhunen--Loeve Basis}: The data-driven linear basis is obtained from the empirical covariance matrix $\boldsymbol{\Sigma} = \mathbb{E}[(\tilde{\mathbf{a}} - \bar{\mathbf{a}})(\tilde{\mathbf{a}} - \bar{\mathbf{a}})^\top]$ on the training trajectories. SVD yields eigenvectors $\mathbf{v}_1, \dots, \mathbf{v}_M$ sorted by decreasing eigenvalue $\lambda_i$. The top $K$ principal components form the rows of $\mathbf{B}_{\text{PCA}}$. While DCT treats actuators independently, PCA adapts to empirical cross-axis kinematic correlations across end-effector axes.

\noindent\textbf{Nonlinear Autoencoder (AE)}: We train an MLP Autoencoder with symmetrical bottleneck architecture ($112 \to 64 \to 14 \to 64 \to 112$) using GELU activations to compress chunks into a $14\text{D}$ continuous latent bottleneck. The AE serves as a reconstruction-optimized nonlinear baseline rather than a quantization-aware learned tokenizer.

\subsection{Symmetric Uniform Quantization}
Let $b$ denote the quantization precision with codebook vocabulary size $V = 2^b$ ($V=256$ for nominal 8-bit quantization). Projection coefficients $\mathbf{c} \in \mathbb{R}^K$ are quantized into discrete tokens $\mathbf{z} \in \{0, \dots, V-1\}^K$. Each coefficient $c_k$ is standardized by its training standard deviation $\sigma_{c, k}$, clipped to $[-5, +5]$, and uniformly quantized:
\begin{equation}
\setlength{\abovedisplayskip}{2.5pt}\setlength{\belowdisplayskip}{2.5pt}
    z_k = \left\lfloor \left( \text{clip}\left(\frac{c_k}{\sigma_{c, k}}, -5, 5\right) + 5 \right) \cdot \frac{V - 1}{10} \right\rceil.
\end{equation}
Dequantization maps $z_k$ back to continuous approximation $\hat{c}_k = (z_k \cdot \frac{10}{V-1} - 5) \cdot \sigma_{c, k}$. For linear orthogonal models (PCA and Temporal-DCT), the continuous standardized chunk is recovered via transpose projection $\hat{\tilde{\mathbf{a}}} = \mathbf{B}^\top \hat{\mathbf{c}}$. For the Autoencoder, the continuous latent vector $\hat{\mathbf{h}} = \hat{\mathbf{c}}$ is decoded via its nonlinear decoder network $\hat{\tilde{\mathbf{a}}} = g(\hat{\mathbf{h}})$. Both are finally de-standardized to restore coordinate-scaled action commands $\hat{\mathbf{a}} \in [-1, 1]^6 \times [0, 1]$. All reported reconstruction MSEs are measured post-quantization in this coordinate-scaled action space.

\section{Offline Representation Analysis}
\label{sec:offline}
\enlargethispage{2\baselineskip}

\subsection{Rate--Distortion and Quantization Sensitivity}

We first examine how the representation basis interacts with token budget and quantization precision. Table~\ref{tab:e11_sweep} reveals a clear regime shift rather than a uniformly superior basis. At moderate and high precision, PCA benefits from concentrating trajectory variance into a compact data-adaptive subspace, outperforming Temporal-DCT in 14 of the 20 evaluated configurations. At the nominal setting ($K=14$, 8-bit), this yields an 8.3\% reduction in reconstruction MSE (0.0083 vs. 0.0091).

This advantage reverses under extreme 2-bit quantization, where Temporal-DCT is consistently better across all token budgets. A plausible structural explanation is that Temporal-DCT preserves channel-wise harmonic structure, whereas each PCA component generally mixes multiple action dimensions. Consequently, coarse coefficient errors in PCA can be redistributed across several physical coordinates after inverse projection. These results indicate that a basis that is favored under nominal reconstruction need not remain optimal once quantization noise dominates the error budget.

\begin{table}[t]
\centering
\caption{Token predictability and policy loss on LIBERO-Spatial ($K=14$, $V=256$; lower is better). Policy loss is teacher-forced on seed-42 Task~0.}
\label{tab:predictability}
\vspace{1.5mm}
\resizebox{\columnwidth}{!}{
\begin{tabular}{lccccc}
\toprule
\textbf{Tokenizer} & \textbf{Entropy (bits)} & \textbf{PPL-1} & \textbf{PPL-2} & \textbf{PPL-3} & \textbf{\shortstack{Policy Loss\\(nats/token)}} \\
\midrule
\textbf{PCA}          & 6.665 & 105.61 & 108.20 & 180.81 & 3.513 \\
\textbf{Temporal-DCT} & \textbf{5.919} & \textbf{92.38} & \textbf{90.11} & \textbf{143.37} & \textbf{3.419} \\
\midrule
\textbf{DCT Advantage}& \textbf{$-$0.746 bits} & \textbf{$-$12.5\%} & \textbf{$-$16.7\%} & \textbf{$-$20.7\%} & \textbf{$-$2.7\%} \\
\bottomrule
\end{tabular}
}
\vspace{-3.5mm}
\end{table}

\subsection{Cross-Task and Cross-Dataset Generalization}

We next test whether the learned PCA basis remains advantageous outside the distribution on which it is fitted. In leave-one-task-out evaluation within LIBERO-Spatial, PCA retains a modest advantage, winning 6 of 10 held-out tasks and reducing macro-average MSE by 6.2\%. (For this offline diagnostic, the PCA basis is refitted on the remaining nine tasks in each fold; this is distinct from our downstream policy protocol, where Task~9 is strictly excluded from fitting and training.) This suggests that learned cross-axis structure transfers across related manipulation tasks.

A different pattern emerges under zero-shot cross-dataset transfer. On LIBERO-Object, neither basis consistently dominates. On LIBERO-Goal, however, the preferred representation depends strongly on token budget: Temporal-DCT achieves 22.9--42.2\% lower MSE at $K\in\{7,14\}$, whereas PCA recovers the advantage at $K\in\{21,28\}$ and improves MSE by up to 28.3\% at $K=28$. This crossover suggests that data-driven bases are most effective when sufficient representational capacity is available to capture domain-specific correlations, while a fixed harmonic prior can be more robust in this transfer setting when the representation budget is strongly constrained.

\subsection{Sequence Predictability}

Geometric fidelity is only one aspect of an action tokenizer for an autoregressive policy; the resulting discrete sequence must also be predictable from preceding tokens. We estimate token entropy and $n$-gram perplexity on disjoint episode splits using Lidstone smoothing (Table~\ref{tab:predictability}).

Despite higher nominal reconstruction error, Temporal-DCT yields an easier prediction problem: position-wise entropy is 0.75 bits lower and perplexity drops by 12.5--20.7\% (Table~\ref{tab:predictability}). Teacher-forced policy loss on seed-42 Task~0 follows the same direction (3.419 vs.\ 3.513 nats/token). Thus, a representation may sacrifice geometric fidelity while producing statistically simpler sequences.

\section{Downstream Robotic Policy Evaluation}
\label{sec:experiments}

\subsection{Evaluation Protocol}

We instantiate each tokenizer in the same Qwen3-VL-4B policy~\cite{qwen3vl}. Action IDs ($V=256$) map to dedicated tokens with dual $128\times128$ image inputs and no proprioception. Policies use LoRA~\cite{lora} ($r=64$, $\alpha=128$) on attention projections, trained for 30k steps with AdamW. Each step predicts $K=14$ tokens, decoded to $T=16$ actions. Normalizers and data-driven tokenizers are fitted on Tasks 0--8; Task~9 is strictly excluded. Paired evaluation over 50 initial states per task/seed across 7 runs yields 3,500 rollouts (113.6 H800 GPU-hours). Macro success uncertainty is estimated via 20,000-draw paired bootstrap.

\subsection{Closed-Loop Control}

Table~\ref{tab:main_rollout} shows that PCA's reconstruction advantage does not translate monotonically into control. Temporal-DCT attains 86.2\% mean seen-task success, compared with 83.2\% for PCA. It leads by 6.2 and 5.3 pp for seeds 42 and 43, respectively; the corresponding paired-bootstrap 95\% intervals for PCA$-$DCT are $[-10.7,-1.8]$ and $[-9.3,-1.3]$ pp. Seed~44 reverses the ordering, with PCA ahead by 2.7 pp, but its interval $[-1.8,+7.1]$ includes zero. Hence, the aggregate 3.0-pp DCT advantage is reproducible in direction for two seeds, but not invariant to policy optimization. Given only three independent policy-training seeds, we therefore treat the cross-seed mean gap as descriptive rather than as an inferential claim over optimization randomness.

PCA has a higher three-seed mean on five of nine seen tasks, but DCT gains 18.0 pp on Tasks~3 and 8, producing the higher macro-average. Both tokenizers score 0/50 on held-out Task~9 across all seeds, precluding cross-task comparison and reflecting a general floor effect.

\subsection{Reconstruction--Control Misalignment}

The nonlinear autoencoder provides a second counterexample to reconstruction-driven tokenizer selection. As shown in Table~\ref{tab:ablations}, it reduces post-quantization MSE by 36.7\% relative to PCA and 41.9\% relative to Temporal-DCT. Nevertheless, its 82.7\% seen-task success lies between PCA (80.0\%) and DCT (86.2\%): the reconstruction ranking AE$>$PCA$>$DCT does not match the control ranking DCT$>$AE$>$PCA. Greater representational flexibility therefore improves geometric fidelity without determining the downstream policy ordering.

\begin{table}[t]
\centering
\caption{Tokenizer ablations on LIBERO-Spatial (seed 42, $K=14$). $G_z$ is mean perturbation gain under $|\delta z|=4$; lower is better.}
\label{tab:ablations}
\vspace{1mm}
\resizebox{\columnwidth}{!}{
\begin{tabular}{lcccc}
\toprule
\textbf{Condition} & \textbf{Representation Type} & \textbf{Recon.\ MSE $\downarrow$} & \textbf{$G_z$ $\downarrow$} & \textbf{Seen $\uparrow$} \\
\midrule
\textbf{PCA}          & Learned linear    & 0.00834          & \textbf{0.0336} & 80.0\% \\
\textbf{Temporal-DCT} & Fixed analytical  & 0.00909          & 0.0337          & \textbf{86.2\%} \\
\textbf{Autoencoder}  & Nonlinear neural  & \textbf{0.00528} & 0.0431          & 82.7\% \\
\bottomrule
\end{tabular}
}
\vspace{-3.5mm}
\end{table}

To probe this discrepancy, we measure sensitivity to discrete token perturbations. Let $\hat{\mathbf{a}}(\mathbf{z}) = \mathcal{D}(Q^{-1}(\mathbf{z}))$ denote the decoded action from token vector $\mathbf{z}$, where $\mathcal{D}(\cdot) = \mathbf{B}^\top(\cdot)$ for linear bases (PCA/DCT) and $\mathcal{D}(\cdot) = g(\cdot)$ for the autoencoder. We evaluate the perturbation gain:
\begin{equation}
\setlength{\abovedisplayskip}{2pt}
\setlength{\belowdisplayskip}{2pt}
G_z = \frac{\left\| \hat{\mathbf{a}}(\mathbf{z}+\delta \mathbf{z}) - \hat{\mathbf{a}}(\mathbf{z}) \right\|_2}{\|\delta \mathbf{z}\|_2}.
\end{equation}
We perturb token coordinates by $\delta z_j\in\{\pm1,\pm2,\pm4\}$, clip to valid vocabulary range, and average $G_z$ over chunks, all $K$ coordinates, and both signs in coordinate-scaled action space ($\mathbf{a} \in [-1, 1]^6 \times [0, 1]$). The autoencoder exhibits 26.5--30.3\% larger mean amplification than linear bases; at $|\delta z_j|=4$, $G_z$ reaches 0.0431 vs.\ 0.0336 for PCA/DCT (Table~\ref{tab:ablations}). This diagnostic identifies decoder sensitivity omitted by reconstruction MSE.

\subsection{Synthesis of Representation Trade-offs}

Together, these findings establish a non-monotonic link between rate--distortion quality and control. PCA attains lower nominal reconstruction error among the linear bases, yet DCT produces more predictable sequences, remains robust under coarse quantization, and achieves higher mean seen-task success in our evaluation. The autoencoder minimizes reconstruction MSE but fails to lead control and incurs the highest perturbation gain. Action-tokenizer evaluation should therefore separate \emph{geometric fidelity}, \emph{sequence predictability}, and \emph{decoder stability}.

\section{Limitations}
\label{sec:limitations}

Our study evaluates LIBERO-Spatial with one VLA backbone, three training seeds, and fixed $K=14$ scalar quantization. Morphological diversity, policy scale, learned codebooks, and BPE compression remain open; the observed ordering should not be assumed to transfer unchanged. Perturbation gain is local rather than causal, and the Task~9 floor precludes held-out conclusions.

\section{Conclusion}
\label{sec:conclusion}

In this controlled autoregressive VLA study, rankings diverge: DCT leads PCA by 3.0 pp in mean seen-task success despite worse reconstruction, while the autoencoder minimizes distortion without leading control. Reconstruction alone therefore cannot select a policy representation; tokenizers require sequence-aware diagnostics and paired closed-loop evaluation.

\clearpage
\balance
\bibliographystyle{IEEEbib}
\bibliography{refs}

\end{document}